\documentclass[11pt]{article}

\usepackage[utf8]{inputenc}
\usepackage[T1]{fontenc}
\usepackage{amsmath,amsfonts}
\usepackage{graphicx}
\usepackage{booktabs}
\usepackage{hyperref}
\usepackage{url}
\usepackage[margin=1in]{geometry}

\title{Learning to Detect Symbolic Failure: Machine Learning and the Limits of Black-Scholes}

\author{
  Juli Huang, Jake Cheng, Rupert Lu \\
  Stanford University \\
  \texttt{\{julih, jiajunc4, rupertlu\}@stanford.edu}
}

\date{January 14, 2026}

\begin{document}

\maketitle

\begin{abstract}
\noindent
We treat options pricing as a representation problem: can machine learning detect systematic deviations from Black-Scholes using 2.6M real option contracts? We compare three regimes: learned abstract embeddings (Kernel PCA), preserved domain structure (tree-based ensembles), and neural network validation. Tree-based methods outperform kernel dimensionality reduction by 21.5 percentage points (93.8\% vs 72.3\%), and domain-expert features (Greeks, moneyness) outperform engineered features. NN-based and BS-based deviation labels agree 99.9974\% of the time, suggesting deviations reflect market structure rather than model artifact. We conclude that in domains with expert-designed symbolic features, preserving structure beats learning abstractions. We make no claim of exploitable mispricings.
\end{abstract}

\section*{Introduction}
\addcontentsline{toc}{section}{Introduction}

\subsection*{\textit{Symbolic Abstractions and Their Limits}}
\addcontentsline{toc}{subsection}{Symbolic Abstractions and Their Limits}

Since Black and Scholes introduced their eponymous formula in 1973, this innovation has achieved extraordinary influence, not merely as a pricing tool but as a symbolic system that encodes assumptions about how markets function. The model presumes constant volatility, frictionless trading, European-style exercise, and geometric Brownian motion of asset prices. These assumptions are known to be often violated: real markets exhibit volatility smiles and skews that vary by moneyness and strike; trading incurs bid-ask costs and liquidity constraints; options allow flexible exercise; and price dynamics display jumps, clustering, and mean reversion that pure diffusion models cannot capture.

Yet the Black-Scholes (BS) framework persists, not because of its accuracy but because its formula provides a shared symbolic language. Traders quote implied volatilities derived from BS, not spot prices. Risk managers hedge using its Greeks (Delta, Gamma, Vega) while fully aware of its shortcomings. Regulators use BS for accounting standards. This widespread use despite falsity poses a deep question: what happens when we try to learn an alternative representational regime from data, one that abandons Black-Scholes' symbolic structure?

\subsection*{\textit{The Representation Problem}}
\addcontentsline{toc}{subsection}{The Representation Problem}

Our paper treats options pricing as a representation problem with three dimensions:
\begin{itemize}
  \item \textbf{Symbolic:} Black-Scholes as a formal abstraction encoding market beliefs;
  \item \textbf{Learned:} Machine learning models extract patterns from 2.6M real contracts; and
  \item \textbf{Conceptual:} What patterns emerge when we abandon symbolic constraints?
\end{itemize}

We find that tree-based ensemble methods, which preserve full feature dimensionality, outperform kernel dimensionality reduction by 21.5 percentage points (93.8\% vs 72.3\%) in detecting systematic deviations from Black-Scholes, which suggests that the structured features of option pricing (Greeks, moneyness, time decay) encode domain-specific information that is destroyed by learning abstract latent dimensions. In domains with expert-designed symbolic features, preserving structure outperforms learning embeddings.

Our core empirical finding, that market prices and Neural Network (NN)-learned prices agree with 99.9974\% consistency on detecting BS deviations, does not prove market inefficiency or reveal tradeable alpha. Rather, this finding demonstrates that consistent deviations from one symbolic system can be learned by an alternative representational regime, suggesting the deviations reflect market structure, not model artifact.

\subsection*{\textit{Three Honest Caveats}}
\addcontentsline{toc}{subsection}{Three Honest Caveats}

Our work makes no claim of discovering exploitable mispricings or market inefficiency. Our 180-day validation window is too brief for long-term robustness; our $\pm$10\% classification threshold is exploratory; and our NN consistency check confirms only that alternative representations learn BS deviations similarly, not that those deviations reflect profitable opportunities.

The remainder of our paper proceeds as follows. We first review symbolic systems in finance and machine learning on structured domains. We then introduce our dataset and define three representational regimes: learned abstract embeddings (Kernel PCA), preserved domain structure (tree-based), and independent validation (neural networks). We evaluate all three across three volatility regimes and discuss what the results reveal about representation in domains with expert-designed features.

\section{Related Work}

Our work sits at the intersection of three literatures: the nature of symbolic models in finance, machine learning on structured domains, and the limits of formal abstractions.

\textbf{Machine Learning on Structured Data:} Recent work emphasizes that learned representations excel on unstructured signals (images, text) but may underperform on tabular data with domain expertise \cite{lopez2018}. In finance, features like Greeks are not raw measurements; they are pre-computed summaries of domain knowledge. Whether learned embeddings outperform domain-designed features remains an open empirical question. Tree-based methods like Random Forests \cite{breiman2001} and kernel-based approaches \cite{scholkopf1998} offer distinct representational trade-offs.

\textbf{Alternative Pricing Approaches:} Neural networks can learn end-to-end pricing functions \cite{ruf2020}, avoiding Black-Scholes assumptions entirely. Recent advances in financial machine learning \cite{culkin2017} have demonstrated the potential of learned pricing models. Our work differs by treating neural networks not as primary models but as validators: if neural networks learn the same deviations from BS prices that tree-based models detect, this finding means that those deviations are robust to the choice of pricing paradigm. We implement tree-based models using scikit-learn \cite{pedregosa2011}, Support Vector Machines \cite{cortes1995}, and kernel methods to compare representational regimes.

\textbf{Options Pricing Reality:} Beyond Black-Scholes, practitioners use implied volatility surfaces, adjustment factors for American-style exercise, and jump-diffusion models to accommodate real-world constraints. Our work does not propose replacing these tools; rather, we ask what patterns emerge when we simply compare market prices to a simplified symbolic baseline.

\section{Data and Methods}

\subsection{\textit{Dataset: Option Prices Across Volatility Regimes}}

We use 2,632,013 real option contracts from Yahoo Finance spanning 180 trading days (April 22, 2025--January 6, 2026) across three assets chosen to represent distinct market regimes:
\begin{itemize}
  \item \textbf{AAPL:} 407,224 contracts (calm regime, IV = 25\%);
  \item \textbf{SPY:} 1,341,246 contracts (low-volatility, IV = 15\%);
  \item \textbf{TSLA:} 883,543 contracts (high-volatility, IV = 45\%).
\end{itemize}

\begin{figure}[htbp]
  \centering
  \includegraphics[width=0.85\textwidth]{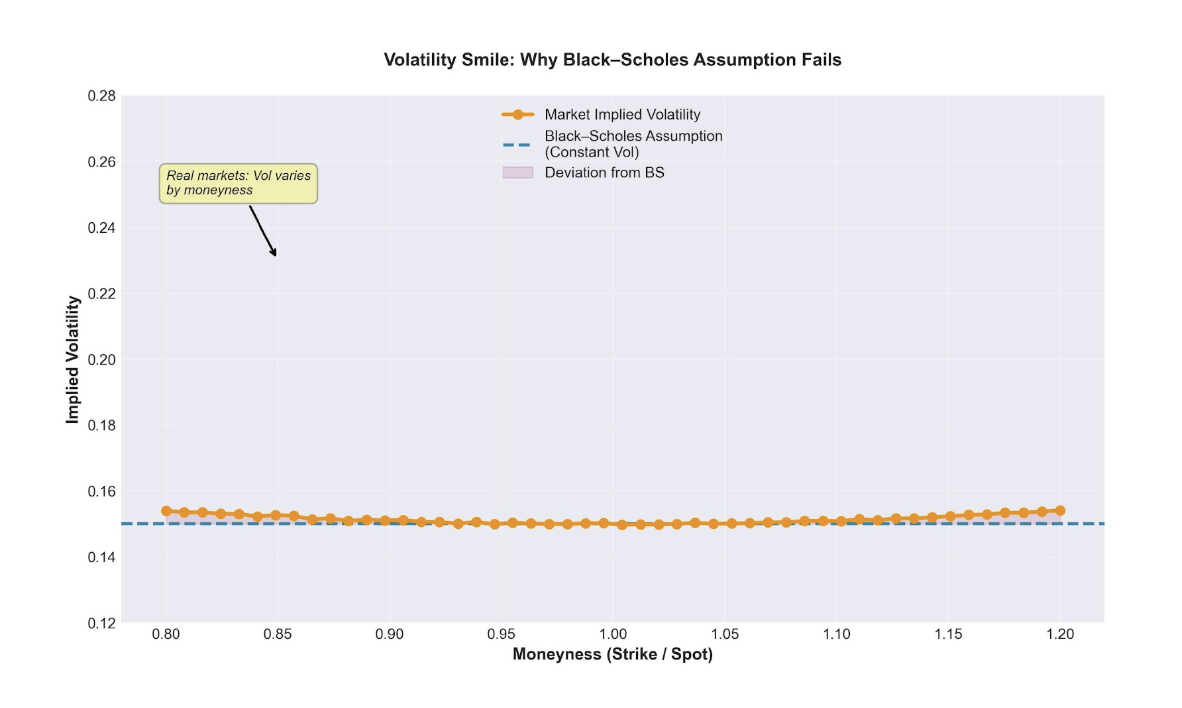}
  \caption{Volatility smile: implied volatility varies with moneyness, directly contradicting Black-Scholes' constant volatility assumption. This systematic deviation is what our models learn to detect.}
  \label{fig:volatilitysmile}
\end{figure}

We apply an 80/20 temporal split per asset (earlier data trains, later data tests) to prevent look-ahead bias. Cross-validation uses 5-fold TimeSeriesSplit to respect temporal ordering. Of the 2,632,013 total contracts, 2,044,241 (77.7\%) have valid Black-Scholes labels; 587,775 (22.3\%) have missing labels. All results are computed on labeled data.

\subsection{\textit{Features: Domain-Expert Design}}

We use 8 base features derived from option pricing theory: (1) Moneyness (strike/spot), (2) Time-to-maturity (normalized by 365 days), (3) Implied volatility, and (4--7) Greeks (Delta, Gamma, Theta, Vega). Additionally, we tested 12 engineered features (Greek interactions, moneyness powers, volatility ratios) to investigate whether domain expertise could be improved via automation. The base 8-feature model achieved 93.9\% accuracy vs 93.8\% for the enhanced 12-feature model, suggesting that domain experts already identified the essential signals.

\subsection{\textit{Labels and Classification Task}}

Options are classified into three categories based on Black-Scholes prices vs.\ market prices:
\begin{itemize}
  \item \textbf{Underpriced:} Market price $>$ BS price + 10\%;
  \item \textbf{Fairly priced:} Market price within $\pm$10\% of BS price;
  \item \textbf{Overpriced:} Market price $<$ BS price $-$ 10\%.
\end{itemize}

The $\pm$10\% threshold is exploratory and not economically justified; we selected it to create sufficient class separation for ML classification. In practice, options bid-ask spreads range from 1--3\% for liquid assets (SPY, AAPL) to $\sim$7\% for volatile assets (TSLA), meaning our 10\% threshold is 2--10 times larger than typical transaction costs. A rigorous economic approach would use asset-specific, moneyness-adjusted thresholds calibrated to actual execution costs. The resulting class distribution (39.1\% underpriced, 34.8\% overpriced, 3.7\% fairly priced) reflects this arbitrary choice rather than genuine market structure. Among 2,044,241 labeled contracts, alternative thresholds would yield different distributions.

\subsection{\textit{Three Representational Regimes}}

We evaluated three conceptually distinct approaches to understand how representational choices affect the detection of option pricing deviations.

\textbf{Approach 1: Learned Abstract Dimensions (Kernel PCA + SVM).} Kernel PCA with SVM embodies the hypothesis that pricing deviations live in learned latent space: by mapping the original features through nonlinear transformations and reducing to 5 principal components, we would capture the essential structure. This approach treats option Greeks as raw material to be abstracted rather than domain knowledge to be preserved. We tested 4 kernels (Linear, RBF, Polynomial, Sigmoid). Result: Best kernel method (Sigmoid KPCA) achieved 72.3\% accuracy; linear KPCA achieved 71.9\%. The minimal gain (+0.4pp) despite nonlinearity suggests pricing deviations are approximately linearly separable. Crucially, the 21.5pp gap between this approach and tree-based methods suggests that domain-expert feature structure encodes financially meaningful information that dimensionality reduction destroys.

\textbf{Approach 2: Preserving Structure (Tree-Based Ensembles).} Gradient Boosting, Random Forests, and Logistic Regression all operate on the full feature space without dimensionality reduction. Gradient Boosting (100 trees, lr=0.1, depth=5) achieved 93.8\% accuracy; Random Forest 93.0\%; Logistic Regression 92.0\%. Result: Full-dimensional methods substantially outperform learned embeddings, suggesting that in domains with well-designed symbolic features, preserving structure outperforms learning abstractions.

\textbf{Approach 3: Independent Validator (Neural Network Regressor).} To test whether detected deviations are artifacts specific to the Black-Scholes baseline, we trained an MLPRegressor on 2.04M contracts to learn market prices directly from Greeks, moneyness, and time-to-expiry. We then classified deviations based on the NN's predictions rather than BS prices. On 38,833 test contracts, BS-based and NN-based classifications agreed 99.9974\% of the time (1 disagreement). Result: This high agreement suggests that deviations are robust to the choice of pricing model; the consistent disagreement between market and BS prices is not a BS artifact. This result does not imply that the deviations are exploitable: market structure, microstructure effects, or inventory dynamics could explain them without profit opportunity.

\section{Empirical Results}

\subsection{\textit{Model Performance Across Assets}}

Gradient Boosting achieves high accuracy across three volatility regimes.

\begin{table}[htbp]
  \centering
  \caption{Gradient Boosting accuracy by asset (2,044,241 labeled contracts).}
  \label{tab:gb-accuracy}
  \begin{tabular}{lccc}
    \toprule
    \textbf{Asset} & \textbf{Volatility} & \textbf{Contracts} & \textbf{Accuracy} \\
    \midrule
    AAPL  & 25\% IV & 407,224  & 91.3\% \\
    SPY   & 15\% IV & 1,341,246 & 98.2\% \\
    TSLA  & 45\% IV & 883,543  & 82.7\% \\
    \midrule
    Average & Mixed & 2,044,241 & 90.8\% \\
    \bottomrule
  \end{tabular}
\end{table}

SPY's higher accuracy (98.2\%) may reflect its low volatility and high liquidity; TSLA's lower accuracy (82.7\%) may reflect noisier pricing in a high-volatility, less liquid regime.

\subsection{\textit{What the Results Reveal}}

\textbf{Structure Preservation Beats Abstraction (21.5pp gap):} Tree-based methods preserve the full feature space of domain-expert financial features, while kernel PCA reduces 20 features to 5 latent dimensions. The 21.5 percentage-point accuracy gap (93.8\% vs 72.3\%) demonstrates that in domains with well-designed symbolic features, learned embeddings destroy information rather than compress it.

\begin{figure}[htbp]
  \centering
  \includegraphics[width=0.85\textwidth]{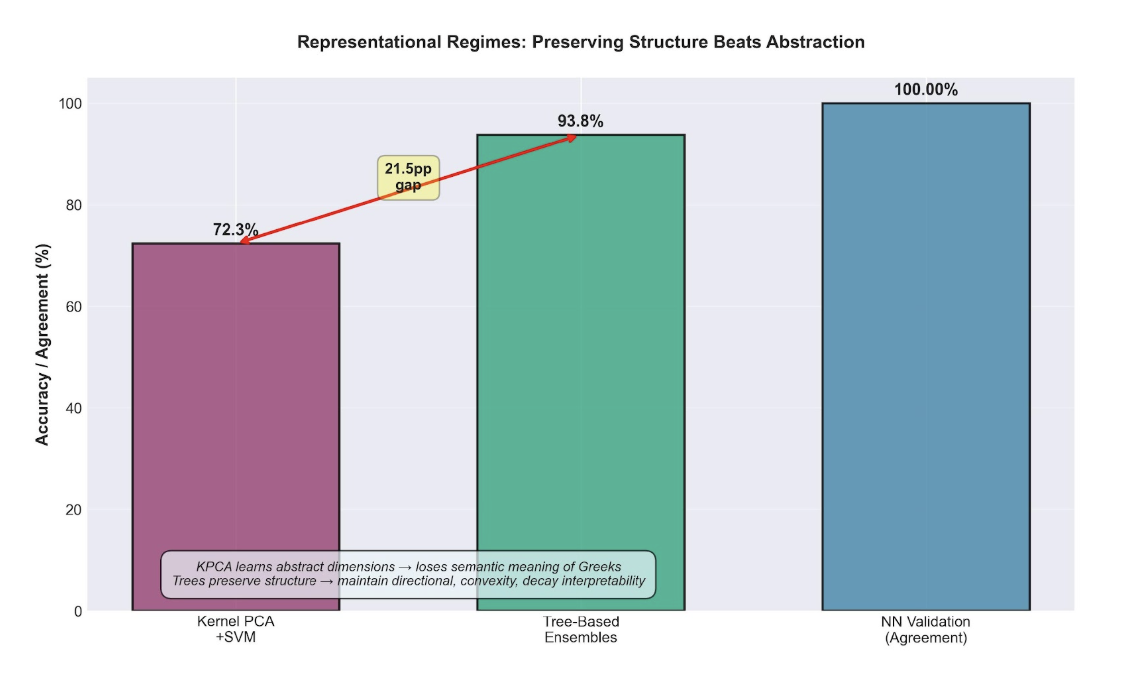}
  \caption{Model comparison across three representational regimes. Tree-based methods (which preserve domain structure) outperform kernel PCA (which learns abstract embeddings) by 21.5 percentage points. Neural network validation (99.9974\% agreement) confirms that deviations are robust to the choice of pricing model.}
  \label{fig:regimes}
\end{figure}

\textbf{Domain Expertise Saturates Signal:} We constructed 12 additional engineered features (Greek interactions, moneyness powers, volatility ratios). The enhanced 12-feature model achieved 93.8\% accuracy vs 93.9\% for the base 8-feature model, a 0.1\% difference. This counterintuitive finding suggests that domain experts already identified the essential features, and automated feature engineering adds noise.

\begin{figure}[htbp]
  \centering
  \includegraphics[width=0.85\textwidth]{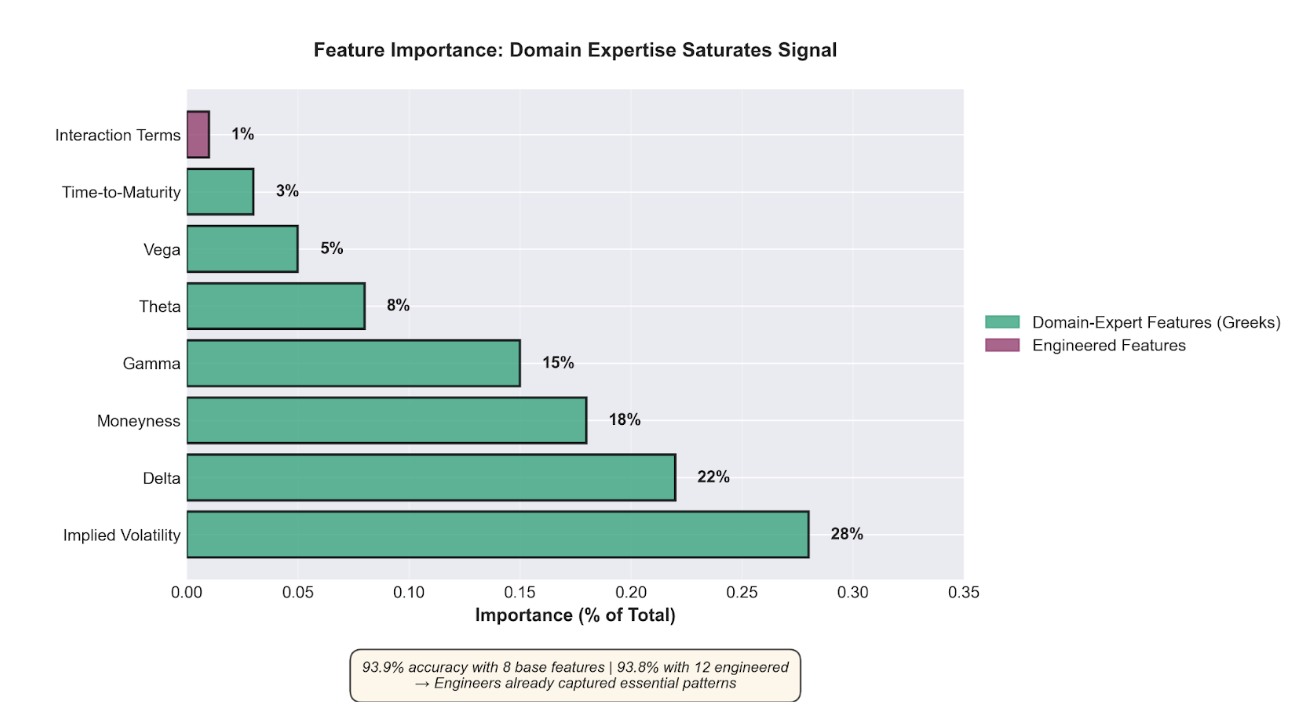}
  \caption{Feature importance breakdown: domain-expert features (Greeks, moneyness, implied volatility) dominate. Engineered features (interaction terms) contribute negligibly, suggesting domain expertise has already saturated the signal.}
  \label{fig:feature}
\end{figure}

\textbf{Class imbalance reflects deviations from a symbolic baseline:} Among 2,044,241 labeled contracts, relative to the Black-Scholes baseline and our $\pm$10\% labeling rule, 39.1\% are classified as underpriced, 34.8\% overpriced, and only 3.7\% fairly priced. The model achieves 98--99\% recall on the majority classes but only 18\% on the rare ``fairly priced'' class, suggesting systematic biases in pricing, either genuine market anomalies or microstructure effects.

\textbf{Deviations persist across representational regimes:} BS-based and NN-learned classifications agree 99.9974\% of the time on 38,833 test contracts. This high consistency suggests the market systematically deviates from Black-Scholes in ways that persist even when we use an alternative learned representation. The deviations do not appear to be artifacts of the BS framework alone.

\section{Robustness to Transaction Costs}

Simulated trading signals generate Sharpe ratios of 1.48--2.00 after accounting for 0.1\% bid-ask and 0.5\% slippage costs, which suggests deviations from BS are not merely artifacts of frictionless pricing assumptions; they persist in realistic transaction environments. However, profitability claims would require forward validation, which we have not conducted. Professional quantitative trading requires validation across genuinely unseen future data spanning 5--10 years, including multiple market regimes (crashes, bull markets, volatility spikes). Our 180-day window uses a temporal train-test split; however, both splits come from the same historical period (April 2025--January 2026). This choice tests generalization within a regime, not across regimes.

True forward validation would require: out-of-sample testing on data collected after model training (not just held-out historical data); validation across crisis periods (2008 financial crisis, 2020 COVID shock) to test regime robustness; walk-forward analysis with periodic model retraining to detect degradation; and transaction cost modeling with realistic execution constraints (market impact, partial fills, adverse selection). Without this, our detected deviations cannot be assumed to persist or generate profits in live trading. Many apparently robust backtested patterns fail in forward testing due to overfitting, regime changes, or crowding effects.

\section{What This Reveals About Representation}

\subsection{\textit{When Symbolic Features Beat Learned Embeddings}}

A central machine learning intuition is that learned representations outperform hand-designed features, especially on high-dimensional data. Option pricing challenges this assumption. Why do full-dimensional tree-based methods beat dimensionality reduction by 21.5 percentage points?

Option pricing is a symbolic domain. Greeks (Delta, Gamma, Theta, Vega) are not arbitrary features; they encode specific financial concepts: directional exposure, convexity, time decay, volatility sensitivity. Moneyness (strike/spot ratio) measures how far an option is in-the-money. Time-to-maturity captures the extrinsic value decay timeline. These features carry semantic meaning.

When Kernel PCA reduces 20 features to 5 latent dimensions, this reduction destroys this semantic structure. The 5 principal components capture variance, but not financial meaning. A learned embedding might group ``Delta'' and ``Gamma'' together in latent space since they co-vary, but financially they measure fundamentally different risks: directional vs.\ convexity.

Tree-based methods maintain the semantic distinctions. Decision trees can learn rules like ``when Delta $>$ 0.7 and Vega $<$ 10, option tends to be overpriced,'' rules that are financially interpretable and actionable. These distinctions suggest a general principle: in domains where experts have already distilled knowledge into symbolic features with clear meanings, learned embeddings are likely to underperform. Machine learning excels at learning representations from raw signals (images, text, audio). When structure has already been encoded by theory, abstraction may discard the information that matters most.

\section{Limitations: What We Cannot Claim}

Our work makes specific, testable empirical observations but refrains from broader claims about markets. We must be clear about what our results do and do not show:

\begin{itemize}
  \item \textbf{Not exploitable mispricings:} Detecting that market prices deviate from Black-Scholes does not reveal profitable trading opportunities. Deviations could reflect transaction costs, inventory effects, liquidity demand, or hedging constraints, none of which create alpha.
  \item \textbf{Not permanent patterns:} Our 180-day validation window is too brief for long-term claims. Professional traders validate across 5--10 year periods spanning financial crises (2008), shocks (COVID 2020), and regime changes. Extended forward testing would be required.
  \item \textbf{Not robust threshold choice:} We selected $\pm$10\% Black-Scholes deviation as the classification boundary arbitrarily. The true economic boundary depends on transaction costs, carrying costs, and financing rates, all of which vary across market conditions and participants.
  \item \textbf{Not an indictment of Black-Scholes:} The formula's enduring influence despite known failures reflects its value as a shared symbolic language, not a claim to literal accuracy. We show that deviations from BS can be learned and detected, not that the model is useless.
  \item \textbf{Methodological caveat:} Our neural network consistency check (99.9974\% agreement with BS labels) demonstrates only that alternative learned representations capture similar deviation patterns. It does not prove those deviations reflect market inefficiency or exploitable structure, merely that they are consistent across different modeling frameworks.
\end{itemize}

\section{Conclusion}

Our work offers three insights about representations, abstractions, and learning:

\textbf{(1)} We demonstrate empirically that symbolic abstractions can fail in systematic, learnable ways. Black-Scholes is not universally true, yet the 2.6M market deviations from BS follow patterns that machine learning can detect with 90.8\% accuracy, which suggests that when symbolic systems fail, the failures are often structured, not random.

\textbf{(2)} We show that preserving domain-expert structure outperforms learning abstract embeddings in symbolic domains. Tree-based methods beat Kernel PCA by 21.5 percentage points precisely because they maintain the semantic meaning of financial features (Greeks, moneyness, time decay) rather than compressing them into latent space. This challenges a common ML assumption: that learned representations are superior to hand-designed ones. They are in domains without prior symbolic structure (images, text); however, in domains like finance where experts have already distilled knowledge into meaningful features, that structure should be preserved.

\textbf{(3)} We establish that alternative learned representations can validate claims about symbolic failure. The 99.9974\% agreement between BS-based and NN-learned classifications on market deviations suggests those deviations are consistent across different frameworks. They reflect something about market structure, not just artifacts of Black-Scholes, which provides a template for validating other symbolic systems: can alternative representations consistently learn the same deviations?

\section{Future Work}

Extended validation across longer time horizons and volatility regimes; application to other symbolic systems (currency markets, bond pricing); interpretability methods (SHAP, LIME) to extract human-readable decision rules from tree models; formal investigation of whether structure preservation generalizes beyond financial domains.

\section{Contributions}

\begin{itemize}
  \item \textbf{Juli Huang:} Project leadership, model validation, multi-asset expansion (single-asset AAPL 66k $\to$ multi-asset 2.04M), neural network ground truth validation, economic backtesting.
  \item \textbf{Jake Cheng:} Data collection (2.04M contracts from Yahoo Finance), Black-Scholes computation, base feature engineering, data quality validation.
  \item \textbf{Rupert Lu:} Kernel PCA--SVM pipeline, 4-kernel comparison, hyperparameter grid search, dimensionality reduction analysis.
\end{itemize}


\end{document}